%% file: main.tex
\pdfoutput=1

\documentclass[11pt]{article}

\usepackage[]{acl}

\usepackage{times}
\usepackage{latexsym}
\usepackage{booktabs}
\usepackage{multirow}
\usepackage{adjustbox}
\usepackage{setspace}

\usepackage{enumitem}
\usepackage{times}
\usepackage{latexsym}
\usepackage{diagbox}

\usepackage[T1]{fontenc}
\usepackage[utf8]{inputenc}
\usepackage{hyperref}

\usepackage{microtype}
\usepackage{colortbl} 

\usepackage{inconsolata}
\usepackage{multirow}
\usepackage{graphicx}
\usepackage{amsmath}
\usepackage{url}   
\usepackage{nicefrac} 
\usepackage{longtable}
\usepackage{bbm}

\usepackage{amssymb}
\usepackage{xcolor}
\usepackage{booktabs}
\usepackage{arydshln}
\usepackage{subcaption}
\usepackage{needspace}
\usepackage{hyperref}
\usepackage{pgfplots}

\definecolor{purple_f}{HTML}{B9B8EF}
\definecolor{purple_b}{HTML}{D4D5FF}
\definecolor{green_f}{HTML}{385723}
\definecolor{green_b}{HTML}{E2F0D9}
\usepackage[most]{tcolorbox}
\tcbset{on line, 
    boxsep=1.0pt, left=0pt, right=0pt,top=0pt,bottom=0pt,
    colframe=white, arc=1pt
}

\usepackage[most]{tcolorbox}

\usepackage{amsmath}
\DeclareUnicodeCharacter{221E}{\ensuremath{\infty}}

\usepackage[T1]{fontenc}

\usepackage[utf8]{inputenc}

\usepackage{microtype}

\usepackage{inconsolata}

\usepackage{graphicx}

\definecolor{arrowgreen}{RGB}{34, 139, 34}
\definecolor{arrowred}{RGB}{178, 34, 34}

\definecolor{gray}{HTML}{D3D3D3} 
\definecolor{green}{HTML}{E2F0D9}
\definecolor{green_f}{HTML}{385723}

\newcommand{\memalpha}{Mem-$\alpha$\xspace}

\usepackage{xspace}

\usepackage{algorithm}
\usepackage{algpseudocode}

\algnewcommand\algorithmicinput{\textbf{Input:}}
\algnewcommand\Input{\item[\algorithmicinput]}
\algnewcommand\algorithmicoutput{\textbf{Output:}}
\algnewcommand\Output{\item[\algorithmicoutput]}

\title{Fine-Mem: Fine-Grained Feedback Alignment for Long-Horizon \\ Memory Management}

\author{
    Weitao Ma$^{1,2}$\thanks{Work done during internship at Meituan.} \quad Xiaocheng Feng$^{1,3}$\footnotemark[2] \quad Lei Huang$^{1}$ \quad Xiachong Feng$^{4}$ \\ 
    \textbf{Zhanyu Ma$^{2}$  \quad Jun Xu$^{2}$\footnotemark[2] \quad Jiuchong Gao$^{2}$\footnotemark[2] \quad Jinghua Hao$^{2}$ \quad Renqing He$^{2}$  \quad Bing Qin$^{1,3}$}\thanks{Corresponding Author.} \\
    $^{1}$ Harbin Institute of Technology, $^{2}$ Meituan \\
    $^{3}$ Peng Cheng Laboratory, $^{4}$ The University of Hong Kong \\
    \texttt{wtma@ir.hit.edu.cn}
}

\begin{document}
\maketitle
\begin{abstract}
Effective memory management is essential for large language model agents to navigate long-horizon tasks.
Recent research has explored using Reinforcement Learning to develop specialized memory manager agents.
However, existing approaches rely on final task performance as the primary reward, which results in severe reward sparsity and ineffective credit assignment, providing insufficient guidance for individual memory operations.
To this end, we propose Fine-Mem, a unified framework designed for fine-grained feedback alignment. 
First, we introduce a Chunk-level Step Reward to provide immediate step-level supervision via auxiliary chunk-specific question answering tasks. 
Second, we devise Evidence-Anchored Reward Attribution to redistribute global rewards by anchoring credit to key memory operations, based on the specific memory items utilized as evidence in reasoning.
Together, these components enable stable policy optimization and align local memory operations with the long-term utility of memory. 
Experiments on Memalpha and MemoryAgentBench demonstrate that Fine-Mem consistently outperforms strong baselines, achieving superior success rates across various sub-tasks.
Further analysis reveals its adaptability and strong generalization capabilities across diverse model configurations and backbones.
\end{abstract}

\section{Introduction}{\label{sec:intro}}
\input{section/Introduction}

\section{Preliminaries}{\label{sec:preliminary}}
\input{section/Preliminaries}

\section{Methodology}\label{sec:methods}
\input{section/Methods}

\section{Experimental Setup}\label{sec:experiments}
\input{section/Experiments}

\section{Results}\label{sec:analysis}
\input{section/Results}

\section{Related Work}\label{sec:related_work}
\input{section/Related_work}

\section{Conclusion}\label{sec:conclu}
\input{section/Conclusion}

\section*{Limitations}\label{sec:limitations}
\input{section/Limitations}

\bibliography{custom}

\appendix
\include{section/Appendix}

\end{document}

%% file: section/Introduction.tex
Large Language Model (LLM) agents have emerged as a powerful paradigm for addressing various downstream tasks, ranging from multi-turn dialogue to complex multi-step reasoning \citep{achiam2023gpt, yang2025qwen3, luo2025large}. 
While LLM agents excel in short-term tasks, they struggle with long-horizon scenarios, which involve evolving goals and information tracking over sessions, far exceeding the capacity of fixed context windows \citep{huang2023survey, liu2025comprehensive}. 
Consequently, robust memory systems are essential to mitigate information decay and ensure the long-term coherence of decision-making \citep{zhang2025survey, wang2024survey, liang2025aimeetsbrainmemory}.

\begin{figure}[t]
\centering
\includegraphics[width=1.0\linewidth]{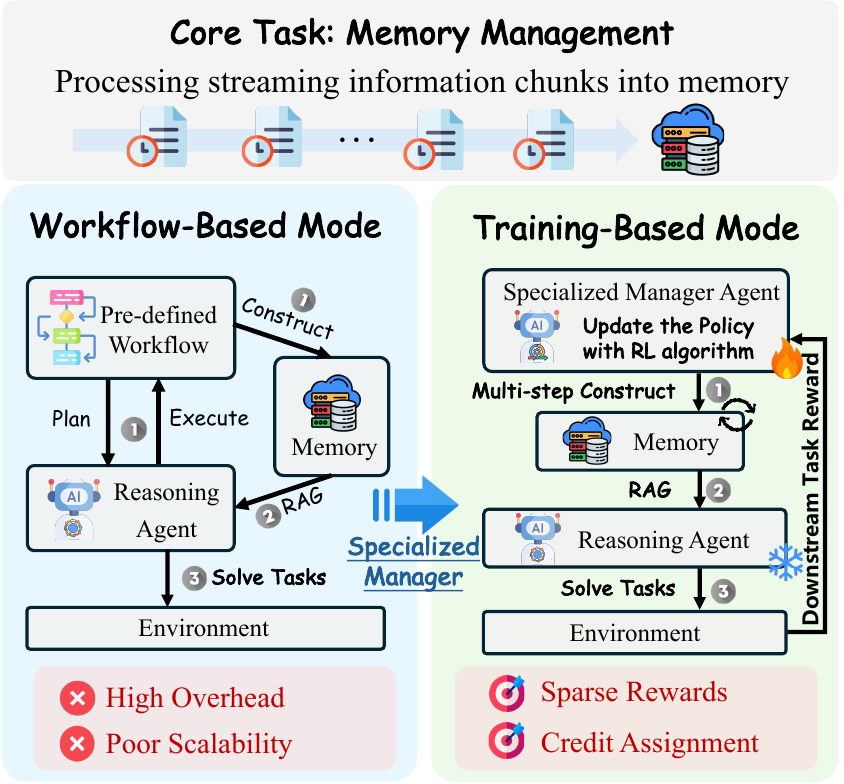}
\caption{Comparison of memory management paradigms. Left: \textit{Workflow-based mode} relies on pre-defined pipelines and strong LLMs, suffering from high overhead and poor scalability. Right: \textit{Training-based mode} utilizes a specialized manager optimized via RL, improving effectiveness but hindered by sparse rewards and ineffective credit assignment problems.
}
\label{fig:comparison_memory_mechanism}
\end{figure}

To bridge this gap, research has shifted from static retrieval to dynamic memory management, focusing on the efficient processing of streaming information chunks into structured memory \citep{wang2025mirix, chhikara2025mem0}. 
As illustrated in Figure~\ref{fig:comparison_memory_mechanism}, current approaches primarily diverge into two paradigms: the \textit{workflow-based mode} and the \textit{training-based mode}.
Workflow-based approaches \citep{xu2025mem, fang2025lightmem}, rely on pre-defined heuristic pipelines and large-scale general LLMs.
Consequently, they often suffer from high computational overhead and poor scalability \citep{hu2025evaluating}.
To mitigate these limitations, recent approaches have been trending towards training-based methods, employing specialized, smaller-scale manager agents optimized through Reinforcement Learning (RL) \citep{yan2025memory, yu2507memagent}.
These agents are typically trained to either fold information in working memory \citep{zhou2025mem1} or manage long-term memory with sequential operations \citep{wang2025mem}. 
However, relying solely on downstream task performance results in \textbf{sparse rewards} and \textbf{ineffective credit assignment}, ultimately undermining the manager's performance in complex tasks.

In this work, we present Fine-Mem, a unified reinforcement learning framework that aligns fine-grained feedback with memory operations. 
To tackle the critical challenges of reward sparsity and ineffective credit assignment, Fine-Mem introduces two components providing dense supervision and principled reward attribution.
Firstly, we propose a \textbf{Chunk-level Step Reward} (CSR) that provides process supervision via chunk-specific question answering. 
By using constructed questions to assess the information retention of individual chunks, CSR yields immediate feedback to mitigate sparse global rewards.
Secondly, we introduce \textbf{Evidence-Anchored Reward Attribution} (EARA) to achieve fine-grained credit assignment. 
By tracing memory items retrieved in downstream tasks back to critical memory operations, EARA maps final utility to specific steps, effectively redistributing the global reward.
Together, these components enable stable policy optimization through Group Relative Policy Optimization, empowering the memory manager to master complex strategies.

To validate Fine-Mem, we conduct extensive experiments on two representative benchmarks, Memalpha and MemoryAgentBench, which assess long-horizon and cross-session memory capabilities. 
Empirical results demonstrate that Fine-Mem consistently outperforms seven competitive baselines, achieving average improvements of 4.4\% on Memalpha and 7.2\% on MemoryAgentBench. 
Moreover, it attains leading performance across all sub-tasks, including Accurate Retrieval, Test-Time Learning, and Long-Range Understanding, while preserving relatively concise memory lengths.
Ablation studies confirm that CSR and EARA jointly enhance task performance while effectively constraining memory length.
Furthermore, extensive evaluations across varied reasoning models and manager backbones highlight the framework's strong generalization ability in diverse settings.

%% file: section/Preliminaries.tex
\subsection{Task Formulation}
The task of memory management can be formulated as a sequential decision-making process over a stream of historical information, denoted as $\mathcal{C} = \{c_1, c_2, \dots, c_T\}$, where each $c_t$ represents a history chunk (e.g., a segment of past dialogue turns). 
Our framework consists of two primary agents: a learnable \textit{Memory Manager} $\pi_\theta$ and a fixed \textit{Reasoning Agent} $\pi_{\text{reason}}$.

At each time step $t$, the \textit{Memory Manager} $\pi_\theta$ observes the current history chunk $c_t$ and the previous memory state $\mathcal{M}_{t-1}$, generating a set of operations $P_t$ to update the memory:
\begin{equation}
    P_t \sim \pi_\theta(\cdot \mid c_t, \mathcal{M}_{t-1}), \quad \mathcal{M}_t = \mathcal{T}(\mathcal{M}_{t-1}, P_t),
\end{equation}
where $\mathcal{T}$ denotes the state transition function that executes the set of operations $P_t$.
The memory state $\mathcal{M}_{t}$ can be viewed as a collection of memory items accumulated up to step $t$.

After processing the entire streaming chunks, the finalized memory state $\mathcal{M}_{T} = \{m_1, m_2, \dots\}$ serves as the knowledge foundation for downstream tasks.
Given a specific query $q_j$, the \textit{Reasoning Agent} $\pi_{\text{reason}}$ first retrieves a relevant subset of memory items and then generates an answer $a_j$ conditioned on both the query and the items:
\begin{equation}
    M_j = \text{Retrieve}(q_j, \mathcal{M}_T),\\
    a_j \sim \pi_{\text{reason}}(\cdot \mid q_j, M_j)
\end{equation}
where $M_j \subseteq \mathcal{M}_T$. This process simulates long-horizon scenarios, where the agent must rely on the accumulated memory to address complex queries.

\begin{figure*}[t]
\centering
\includegraphics[clip, width=1.0\linewidth]{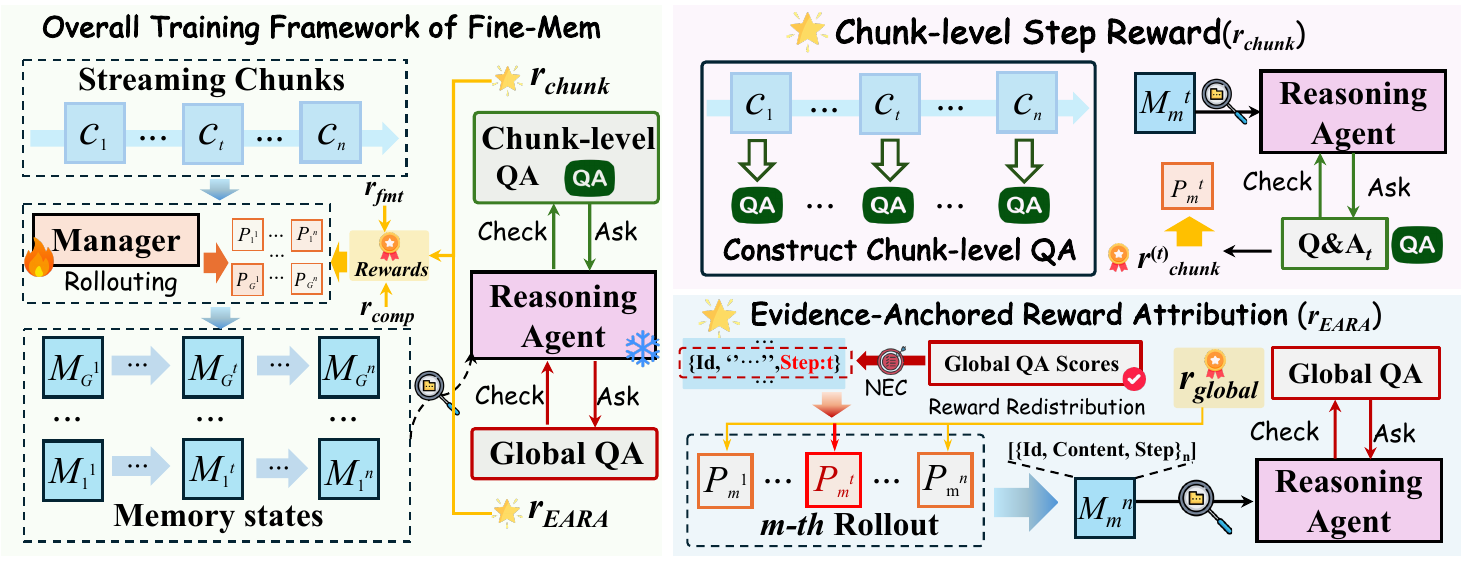}
\caption{An overview of Fine-Mem. Left: The overall training framework. Right: Two core components designed to enhance training: (1) Chunk-level Step Reward (§\ref{meth:chunk_step_reward}), which addresses reward sparsity by generating chunk-level QA tasks to provide step-level feedback for memory operations; (2) Evidence-Anchored Reward Attribution (§\ref{meth:EARA}), which resolves the credit assignment challenge by redistributing global rewards back to specific rollout steps.}
\label{fig:methods}
\end{figure*}

\subsection{Memory Architecture and Operations}
\paragraph{Memory Architecture} 
We adopt a unified single-layer memory architecture to free the \textit{Memory Manager} from deciding where information should be stored.
Specifically, each memory item $m_i \in \mathcal{M}$ is represented as a triplet $\{id_i, content_i, s_i\}$, corresponding to a unique identifier, the stored content, and the specific update step, respectively. 
More complex multi-layer memory structures would impose substantial challenges on the \textit{Memory Manager}, as they typically require carefully designed supervision or reward signals to guide memory placement.
However, it isn't easy to specify in practice and may even degrade the manager’s effectiveness when misaligned.

\paragraph{Operation Space}
Following prior work \citep{wang2025mem, yan2025memory}, the \textit{Memory Manager} operates over a compact operation space consisting of four atomic operations: \textbf{INSERT}, \textbf{UPDATE}, \textbf{DELETE}, and \textbf{SKIP}. 
This minimal set affords flexible memory manipulation by enabling addition, refinement, removal, and selective retention of information, while allowing complex behaviors to emerge through the composition of these simple primitives. 
A more detailed definition of the operations is provided in Appendix~\ref{app:operation_definitions}.

%% file: section/Methods.tex
In this section, we present Fine-Mem, a unified reinforcement learning framework for training the memory manager with fine-grained reward assignment, as illustrated in Figure \ref{fig:methods}.
Fine-Mem comprises two key components: (1) \textbf{Chunk-level Step Reward}, which introduces chunk-level QA to provide richer step-level supervision for operations, and (2) \textbf{Evidence-Anchored Reward Attribution}, which redistributes global rewards back to individual rollout steps for more precise credit assignment.

\subsection{Chunk-level Step Reward}
\label{meth:chunk_step_reward}
Current reinforcement learning methods for memory managers rely primarily on sparse, rollout-level rewards, providing no direct supervision for individual decision steps.
However, unlike traditional long-horizon reasoning tasks (e.g., mathematical reasoning or code generation), memory management can be viewed as a sequence of locally grounded decisions triggered by incoming chunks, where each step mainly depends on the current observation and the existing memory state.

An ideal reward for each step-level operation set should jointly consider two aspects: the quality of storing information within the current chunk, and its influence on the handling of information from other chunks.
Accordingly, we introduce chunk-level step rewards ($r_{\textit{chunk}}$), which are computed on the chunk-level QA pairs to model local information processing, while retaining global QA accuracy to capture cross-chunk effects.

The chunk-level QA pairs are generated by first extracting concise, factoid-style questions from each chunk using \texttt{GPT-4o-mini}, with paraphrased formulations included to enhance generalization.
To ensure that each QA pair is fully grounded in the chunk, a verifier model answers each question using only the corresponding chunk content, and any pair that cannot be correctly answered is discarded. 
After removing duplicates, a fixed budget of five QA pairs is retained per chunk.
The complete construction process is detailed in the Appendix~\ref{app:chunk_qa}.
Together, this dual QA accuracy design encourages precise acquisition of new information while ensuring coherent integration across the entire sequence.

\input{section/tables/eara_pseudocode}

\subsection{Evidence-Anchored Reward Attribution}
\label{meth:EARA}
Although dense chunk-level step rewards capture the local quality of individual memory operations and alleviate reward sparsity, they remain insufficient to reflect the long-term utility of memory management across an entire chunk stream. 
To reasonably attribute the global QA signal to the memory operations that truly matter, we introduce \textbf{Evidence-Anchored Reward Attribution (EARA)}, a mechanism that bridges the final global reward ($r_{\textit{global}}$) and intermediate memory actions by tracing downstream reasoning back to the specific memory operations that enabled them. 
EARA is instantiated through two complementary mechanisms, which is summarized in Algorithm~\ref{alg:eara_memory}.

We first quantify the objective utility of each memory operation step based on its evidential support for downstream reasoning. 
Consider a rollout consisting of $T$ memory steps and evaluated by $n$ global QA queries. 
Let $s_j \in [0,1]$ denote the score obtained for question $q_j$, and let $M_j$ be the set of memory entries retrieved by the reasoning agent to produce its answer.
We define the \textit{Normalized Evidence Contribution (NEC)} of step $t$ as:
\begin{equation}
N_t = \sum_{j=1}^{n} \sum_{\substack{m \in M_j \\ \phi(m)=t}} \frac{s_j}{|M_j| \cdot n},
\label{eq:nec}
\end{equation}
where $\phi(m)$ identifies the memory operation step from which memory item $m$ originates. 

To prevent over-sensitivity to individual evaluation questions and to further stabilize gradient estimation, we adopt a soft reward redistribution strategy. 
Specifically, the final EARA reward assigned to step $t$ is computed as a combination of a uniform baseline reward and the evidence-driven contribution:
\begin{equation}
r_{\textit{EARA}}^{(t)} = (1-\beta)\frac{r_{\textit{global}}}{T} + \beta N_t, 
\label{eq:eara_final}
\end{equation}
where $\beta \in [0,1]$ is an attribution factor controlling the strength of evidence-based credit assignment. 
The uniform term provides a \emph{participation credit} to all steps in the rollouts, ensuring a stable learning signal for exploration, while the NEC term assigns \emph{performance-based credit} to the specific memory operations that produced effective evidence. 
Importantly, EARA redistributes the full global reward of each rollout across all memory steps, as shown in Appendix~\ref{app:proof_eara}, ensuring the attribution remains consistent with the original optimization objective.

\subsection{Policy Optimization via GRPO}
Beyond correctness from $r_\textit{chunk}$ and $r_\textit{EARA}$, manager operations are also encouraged to be efficient in terms of memory usage and comply with the formatting required for deployment. 
To this end, we incorporate auxiliary rewad for invalid function formatting ($r_\textit{fmt}$) and memory compression ($r_\textit{comp}$), which have been shown effective in prior work \citep{wang2025mem}. 
The total reward $r_t$ for a set of operations $P_{t}$ is formulated as a weighted sum:
\begin{equation}
r_t = r_\textit{EARA}^{(t)} + r_\textit{fmt}^{(t)} + w_1 r_\textit{chunk}^{(t)} + w_2 r_\textit{comp}
\label{eq:reward}
\end{equation}
where $w_1$ and $w_2$ are balancing hyperparameters for the auxiliary rewards, each normalized to lie in the range $[0,1]$.
Formal definitions of all reward components are provided in Appendix~\ref{app:hybrid_reward}.

We then optimize the policy using GRPO \citep{shao2024deepseekmath}.
At each step $t$, the simplified objective with memory operations is:
\begin{equation}
\mathcal{L}_{\text{policy}} \approx \sum_{G} \log \rho_\theta(P_t \mid \mathcal{M}_{t-1}, c_t) \cdot A_t,
\end{equation}
where $G$ represents the number of rollouts, $P_t$ denotes the operations, $A_t$ is the corresponding advantage and $\rho_\theta(\cdot)$ is the corresponding importance ratio. 
The complete GRPO objective is introduced in Appendix~\ref{appendix:grpo}.

\input{section/tables/main_memalpha}

%% file: section/tables/eara_pseudocode.tex
\begin{algorithm}[t]
\caption{Evidence-Anchored Reward Attribution for a Single Rollout}
\label{alg:eara_memory}
\begin{algorithmic}[1]

\Input Global QA scores $\{s_j\}_{j=1}^{n}$, retrieved memory sets $\{M_j\}$,
memory-to-step map $\phi(\cdot)$, global reward $r_{\textit{global}}$, factor $\beta$
\Output Step-level rewards $\{r_{\textit{EARA}}^{(t)}\}_{t=1}^{T}$

\State Initialize evidence contributions $N_t \leftarrow 0 \quad \forall t$
\For{$j = 1$ to $n$}
    \For{each memory item $m \in M_j$}
        \State $t \leftarrow \phi(m)$
        \State $N_t \mathrel{+}= \dfrac{s_j}{|M_j| \cdot n}$
    \EndFor
\EndFor


\For{$t = 1$ to $T$}
    \State $r_{\textit{EARA}}^{(t)}
    \leftarrow (1-\beta)\dfrac{r_{\textit{global}}}{T}
    + \beta\,{N_t}$
\EndFor

\State \Return $\{r_{\textit{EARA}}\}$
\end{algorithmic}
\end{algorithm}

%% file: section/tables/main_memalpha.tex
\begin{table*}[t]
    \centering
    \small
    \resizebox{\textwidth}{!}{
    \begin{tabular}{lccccccccc}
    \toprule
        \multirow{3}{*}{\textbf{Method}}
         & \multicolumn{3}{c}{\textbf{AR}} & \multicolumn{3}{c}{\textbf{TTL}} & \textbf{LRU} & \multirow{3}{*}{\textbf{Avg.} $\uparrow$} & \multirow{3}{*}{\textbf{Len.} $\downarrow$} \\
        \cmidrule(lr){2-4} \cmidrule(lr){5-7} \cmidrule(lr){8-8}
        & SQuAD & HotpotQA & PerLTQA & TREC-C & NLU & Pubmed & BookSum & & \\
        \midrule
        \rowcolor{gray!50} 
        \multicolumn{10}{c}{\textbf{\textit{Non-Constructive Memory}}}  \\
        Long-Context 
        & 0.742 & \underline{0.852} & 0.605 & 0.623 & \underline{0.708} & 0.533 & 0.052 & 0.588 & 10.8K  \\
        RAG-Top2
        & 0.762 & 0.849 & 0.623 & 0.612 & 0.508 & \underline{0.570} & 0.042 & 0.567 & 11.3K  \\
        \midrule
        \rowcolor{gray!50} 
        \multicolumn{10}{c}{\textbf{\textit{Workflow-Based Memory Systems}}}
        \\
        A-Mem
        & \underline{0.827} & 0.826 & 0.227 & 0.628 & 0.686 & 0.443 & 0.181 & 0.545 & 11.5K \\
        LightMem
        & 0.214 & 0.271 & 0.283 & \textbf{0.833} & 0.671 & 0.214 & 0.154 & 0.377 & 2.12K \\
        \midrule
        \rowcolor{gray!50} 
        \multicolumn{10}{c}{\textbf{\textit{Train-Based Memory Agents}}}
        \\
        MemAgent
        & 0.091 & 0.140 & 0.052 & 0.562 & 0.290 & 0.343 & 0.103 & 0.226 & \underline{0.84K} \\
        MEM1
        & 0.039 & 0.083 & 0.068 & 0.269 & 0.056 & 0.175 & 0.085 & 0.111 & \textbf{0.17K} \\
        \memalpha
        & 0.786 & 0.832 & \underline{0.659} & 0.666 & 0.658 & 0.545 & \underline{0.187} & \underline{0.619} & 7.90K  \\
        \rowcolor{blue!15} 
        Fine-Mem
        & \textbf{0.836}  & \textbf{0.872} & \textbf{0.693} & \underline{0.675} & \textbf{0.728} & \textbf{0.624}  & \textbf{0.213} & \textbf{0.663} & 10.2K  \\
    \bottomrule
    \end{tabular}}
    \caption{Performance in the memory on validation datasets of Memalpha. \textbf{Bold} and \underline{underline} numbers indicate the best and second performance among evaluated methods. The \textbf{Len.} denotes the average memory length per sample in thousands of tokens.}
    \label{tab:memalpha}
\end{table*}

%% file: section/Experiments.tex
\subsection{Datesets and Evaluation}
\paragraph{Datasets} Following \memalpha \citep{wang2025mem}, we train our model on the Memalpha training corpus, specifically augmented with constructed chunk-level QA pairs. 
We conduct a comprehensive evaluation to assess both in-distribution (ID) performance, using the Memalpha validation set, and out-of-distribution (OOD) generalization, using MemoryAgentBench \citep{hu2025evaluating}, which features significantly longer contexts and higher complexity compared to existing benchmarks like LoCoMo \citep{maharana2024evaluating} and LongMemEval \citep{wu2024longmemeval}.

\paragraph{Metrics}  We assess three core capabilities of memory mechanisms utilizing the corresponding sub-datasets from MemoryAgentBench:
(1) \textbf{Accurate Retrieval (AR)}: This category includes Single-Doc and Multi-Doc measured by \textit{Substring Exact Match}, and LME(S*) measured by \textit{LLM-as-Judge}.
(2) \textbf{Test-Time Learning (TTL)}: Covers five classification domains (TREC-C, NLU, TREC-F, Clinic, Banking77), evaluated via \textit{Classification Accuracy}.
(3) \textbf{Long-Range Understanding (LRU)}: Utilizes InfBench-Sum for summarization tasks.
Detailed descriptions and statistics for all datasets are provided in Appendix~\ref{app:datasets}.

\subsection{Baselines}
To validate the effectiveness of Fine-Mem, we compare it against seven baselines, which can be grouped into three major categories,  with detailed settings provided in Appendix~\ref{app:baselines}:
\begin{itemize}[leftmargin=*]
\item \textbf{Non-Constructive Memory}.
These methods operate without explicit memory construction and rely solely on raw context or simple retrieval mechanisms.
(1) \textbf{Long-Context} strategy directly processes the entire memory context using its maximum window size.

(2) \textbf{RAG-Top2} baseline applies a retrieval-augmented strategy based on BM25 to select the two most relevant chunks, which are then used to answer the query.

\item \textbf{Workflow-Based Memory Systems}.
This category builds external memory modules through LLM-driven workflows.
(1) \textbf{A-Mem} \citep{xu2025mem} is a dynamic agentic memory system that creates, links, and updates structured memories to support cross-session reasoning.
(2) \textbf{LightMem} \citep{fang2025lightmem} is a cognitively-inspired system that optimizes efficiency through compressive sensory filtering, topic-aware consolidation, and decoupled sleep-time update.

\item \textbf{Train-Based Memory Agents}.
These approaches formulate memory management as a learnable decision-making process.
(1) \textbf{MemAgent} \citep{yu2507memagent} is trained to iteratively process all available chunks according to a task description and forms an internal memory state from which answers are produced.
(2) \textbf{MEM1} \citep{zhou2025mem1} functions as an agent that maintains a single paragraph of memory, which it continuously retrieves and updates as new information becomes available.
(3) \textbf{\memalpha} \citep{wang2025mem} employs a GRPO-based training objective to optimize an external memory manager operating over a hierarchical memory system.
\end{itemize}

\input{section/tables/main_memoryagentbench}

\subsection{Implementation Details} 
We implement Fine-Mem by building upon the VERL framework \citep{sheng2024hybridflow}. 
During main experiments, we adopt BM25 for retrieval, utilize \texttt{Qwen3-4B} as the backbone manager model, and deploy a long-context \texttt{Qwen3-32B} model via vLLM \citep{kwon2023efficient} as the reasoning agent. 
We set the learning rate to $1\times 10^{-6}$, the batch size to 32, and \text{grpo\_rollout\_n} to 8, following the settings used in \memalpha. 
The models are trained for 2 epochs, and we report the performance of the final checkpoint.
The reward weights in Eq.\ref{eq:reward} are configured as $w_1 = 0.5, w_2 =0.05, \beta=0.5$. Further implementation details of Fine-Mem, please refer to Appendix \ref{app:fine-mem}.

%% file: section/tables/main_memoryagentbench.tex
\begin{table*}[t]
    \centering
    \small
    \resizebox{\textwidth}{!}{
    \begin{tabular}{lccccccccccc}
    \toprule
        \multirow{3}{*}{\textbf{Method}}
         & \multicolumn{3}{c}{\textbf{AR}} & \multicolumn{5}{c}{\textbf{TTL}} & \textbf{LRU} & \multirow{3}{*}{\textbf{Avg.} $\uparrow$} & \multirow{3}{*}{\textbf{Len.} $\downarrow$}\\
        \cmidrule(lr){2-4} \cmidrule(lr){5-9} \cmidrule(lr){10-10}
        & Single-Doc & Multi-Doc & LME(S) & TREC-C & NLU & TREC-F & Clinic & Banking77 & InfBench & & \\
        \midrule
        \rowcolor{gray!50} 
        \multicolumn{12}{c}{\textbf{\textit{Non-Constructive Memory}}}  \\
        Long-Context
            & 0.280 & 0.270 & 0.292 & 0.640 & \underline{0.740} & 0.340 & \underline{0.860} & 0.770 & 0.125 & 0.480 & 33K \\
        RAG-Top2
           & 0.690 & 0.450 & \textbf{0.581} & \underline{0.690} & 0.650 & 0.210 & 0.700 & 0.750 & 0.065 & 0.532 & 209K \\
        \midrule
        \rowcolor{gray!50} 
        \multicolumn{12}{c}{\textbf{\textit{Workflow-Based Memory Systems}}}
        \\
        A-Mem
           & 0.420  & 0.360 & 0.427 & 0.680 & 0.720 & 0.400 & 0.170 & \underline{0.780} & 0.103 & 0.451 & 210K \\
        LightMem
           & 0.270 & 0.320 & 0.420 & 0.450 & 0.610 & 0.250 & 0.490 & 0.480 & 0.010 & 0.367 & 42.5K \\
        \midrule
        \rowcolor{gray!50} 
        \multicolumn{12}{c}{\textbf{\textit{Train-Based Memory Agents}}}
        \\
        MemAgent
            & 0.070 & 0.160 & 0.050 & 0.370 & 0.260 & 0.210 & 0.250 & 0.370 & 0.043 & 0.198 & \underline{0.92K} \\
        MEM1
            & 0.070 & 0.180 & 0.090 & 0.180 & 0.000 & 0.000 & 0.090 & 0.000 & 0.029 & 0.071 & \textbf{0.21K} \\
        \memalpha
            & \underline{0.740} & \underline{0.680} & 0.520 & \textbf{0.710} & 0.710 & \underline{0.410} & 0.730 & 0.700 & \underline{0.129} & \underline{0.592} & 129K  \\
        \rowcolor{blue!15} 
        Fine-Mem
            & \textbf{0.760} & \textbf{0.720}  & \underline{0.550} & \underline{0.690}  & \textbf{0.800}  & \textbf{0.590}  & \textbf{0.890}  & \textbf{0.820} & \textbf{0.153} & \textbf{0.664} & 148K \\
    \bottomrule
    \end{tabular}}
    \caption{Performance in the memory on MemoryAgentBench. \textbf{Bold} and \underline{underline} numbers indicate the best and second performance among evaluated methods. The \textbf{Len.} denotes the average memory length per sample in thousands of tokens.}
    \label{tab:memoryagentbench}
\end{table*}

%% file: section/Results.tex
\subsection{Main Results}
\label{main_results}
We report the main experimental results in Tables \ref{tab:memalpha} and \ref{tab:memoryagentbench}. 
Fine-Mem achieves superior overall performance, scoring the highest averages of 0.663 on MemAlpha and 0.664 on MemoryAgentBench. 
It consistently attains either the best or second-best results across all sub-capabilities, including Accurate Retrieval, Test-Time Learning, and Long-Range Understanding.

Compared to the state-of-the-art baseline \memalpha, Fine-Mem demonstrates consistent improvements, achieving an average improvement of 4.4\% on Memalpha and 7.2\% on MemoryAgentBench, while maintaining a compact memory length relative to the total input chunks. 
These results validate the effectiveness of incorporating fine-grained supervision into the \textit{memory manager}, suggesting that precise training signals are essential for learning memory management.

In contrast, workflow-based memory agents, despite utilizing strong LLM backbones, suffer from rigid behavioral patterns that limit their generalization to complex tasks. 
For instance, while LightMem maintains a substantial memory length of 42.5K tokens on MemoryAgentBench, its reliance on iterative pre-compression for long chunks leads to information loss, degrading its performance on the Accurate Retrieval subset. 
Meanwhile, train-based methods like Mem1 and MemoryAgent are designed to efficiently fold information of working memory but struggle to adequately scale to external long-term memory management.

Overall, these results demonstrate that \textit{Fine-Mem consistently delivers robust performance for long-term memory management}.

\input{section/tables/ablation_module}

\input{section/tables/ablation_reward_weight_memalpha_new}

\subsection{Ablation Studies}
\label{sec:ab_study}
In this section, we present ablation studies to evaluate the contribution of each Fine-Mem module and the impact of hyperparameters related to the total reward and EARA.

\paragraph{Ablation of Fine-Mem Modules.}
We evaluate three variants of Fine-Mem to isolate the contributions of specific modules: 'OR-Based' (outcome reward only), 'w/ CSR' (adding Chunk-level Step Reward), and 'w/ EARA' (adding Evidence-Anchored Reward Attribution).
As detailed in Table \ref{tab:ablation_module}, incorporating CSR improves the manager's average performance from 0.627 to 0.639. 
However, this gain comes at the cost of a relatively larger memory length. 
Conversely, the EARA mechanism proves highly effective for compression, achieving the lowest average memory length of 60.7K. 
Yet, when used in isolation, its sparse reward signals lead to a performance drop to 0.622.
By combining both modules, Fine-Mem achieves the best overall performance of 0.663 while maintaining a controlled average memory size of 79.1K. 
This result confirms that the two mechanisms are mutually complementary, effectively balancing high retrieval accuracy with memory efficiency.

\input{section/figures/ablation_beta}

\paragraph{Ablation of Total Reward Hyperparameters.}
We investigate the influence of the total reward weights
$w_1$ and $w_2$ on the balance between dense process rewards and compression-based rewards, as outlined in Eq. \ref{eq:reward} and summarized in Table~\ref{tab:ab_rewards}. 
Our findings indicate that a high value of $w_1$ encourages excessive per-step memorization, leading to longer memory lengths, which negatively affects the performance of the memory manager.
Conversely, increasing $w_2$ enhances memory compactness but risks discarding crucial information. 
This leads to performance degradation and, notably, results in longer memory lengths on the out-of-distribution dataset MemoryAgentBench.
Based on these observations, we set $w_1=0.5$ and $w_2=0.05$ as the default configuration, which provides an optimal trade-off between memory length and downstream task performance.

\input{section/figures/ablation_manager_reasoning_memalpha}

\paragraph{Ablation of the Factor in EARA.}
We conduct a comprehensive sensitivity analysis of EARA with respect to the hyperparameter $\beta$, as visualized in Figure~\ref{fig:ablation_beta}. 
We observe that the performance peaks at $\beta = 0.5$, indicating that a balanced reward assignment is essential. 
Specifically, increasing $\beta$ beyond 0.5 leads to significant performance degradation, particularly on the out-of-distribution MemoryAgentBench. 
A high $\beta$ may allocate the most global reward to specific steps, which induces supervisory sparsity and hinders generalization to out-of-distribution tasks.
Conversely, a low $\beta$ setting ignores the contribution of key memory operations, providing insufficient guidance to the manager and thereby hindering the acquisition of effective memory maintenance strategies. 
Consequently, we adopt $\beta = 0.5$ as the default configuration across all main results and ablation studies.

\subsection{Further Analysis}

\paragraph{Generalization of the Memory Manager.}
To evaluate the generalization of memory managers, we benchmark Fine-Mem against baselines (\texttt{Qwen3-4B} and \texttt{GPT-4o-mini}) across different reasoning models, as illustrated in Figure~\ref{fig:ab_manager_reasoning_memalpha} and Figure~\ref{fig:ab_manager_reasoning_MAB}. 
The results demonstrate that Fine-Mem consistently achieves the highest average performance regardless of the reasoning model employed. 
Specifically, it reaches a score of 0.663 with \texttt{Qwen3-32B}, significantly surpassing the baselines, and maintains a similar advantage with \texttt{GPT-4o-mini}. 
This confirms the robustness and effectiveness of Fine-Mem across diverse downstream architectures.

\input{section/tables/ablation_models}

\paragraph{Impact on Different Backbones.}
Furthermore, we assess the effectiveness of Fine-Mem by applying it to specific backbone models, including \texttt{Qwen3-4B}, \texttt{Qwen3-1.7B}, and \texttt{Llama3.2-3B}. 
As shown in Table~\ref{tab:ablation_models}, Fine-Mem yields consistent average performance improvements across these architectures. 
These results further confirm the robustness and general applicability of Fine-Mem, indicating its potential to enhance models of varying scales and types. 

%% file: section/tables/ablation_module.tex
\begin{table}[t]
\centering
\small
\begin{tabular}{llccc}
\toprule
\textbf{Method} & \textbf{Metric} & \textbf{Memalpha} & \textbf{MAB.} & \textbf{Avg.} \\ \midrule

\multirow{2}{*}{OR-Based} & Perf. $\uparrow$  & 0.648 & 0.605 & 0.627 \\
 & Len. $\downarrow$ & 10.7K & 174K  & 92.4K \\ \cmidrule{1-5}

\multirow{2}{*}{\quad w/ \textit{CSR}} & Perf. $\uparrow$ & \underline{0.657} & \underline{0.621} & \underline{0.639} \\
 & Len. $\downarrow$ & 13.4K & 159K  & 86.2K \\ \cmidrule{1-5}

\multirow{2}{*}{\quad w/ \textit{EARA}} & Perf. $\uparrow$ & 0.641 & 0.603 & 0.622 \\
 & Len. $\downarrow$ & \textbf{9.31K} & \textbf{112K}  & \textbf{60.7K} \\ \cmidrule{1-5}

\multirow{2}{*}{Fine-Mem} & Perf. $\uparrow$ & \textbf{0.663} & \textbf{0.664} & \textbf{0.663} \\
 & Len. $\downarrow$ & \underline{10.2K} & \underline{148K}  & \underline{79.1K} \\ \bottomrule
\end{tabular}
    \caption{Ablation study of different components in Fine-Mem on Memalpha and MemoryAgentBench (\textbf{MAB.}) datasets. 
    \textbf{Bold} and \underline{underlined} numbers indicate the best and second results for each metric.}
    \label{tab:ablation_module}
\end{table}

%% file: section/tables/ablation_reward_weight_memalpha_new.tex
\begin{table*}[t]
    \centering
    \small
    \resizebox{\textwidth}{!}{
    \begin{tabular}{llcccccccccc}
    \toprule
        \multirow{3}{*}{$w_1$} &
        \multirow{3}{*}{$w_2$} & 
        \multicolumn{4}{c}{\textbf{Memalpha}} & 
        \multicolumn{4}{c}{\textbf{MemoryAgentBench}} & 
        \multicolumn{2}{c}{\textbf{Avg.}} \\
        \cmidrule(lr){3-6} \cmidrule(lr){7-10} \cmidrule(lr){11-12}
        &  & AR & TTL & LRU & Len. &AR & TTL & LRU & Len. & Perf. $\uparrow$ & Len. $\downarrow$ \\
        \midrule
        0.0 & 0.05
        & 0.767 & 0.662 & 0.202 & 9.31K & 0.463 & \textbf{0.776} & \textbf{0.160} & \textbf{112K} & 0.505 & \textbf{60.7K}  \\
        \rowcolor{green!80}
        0.5 & 0.05
        & \textbf{0.800} & \underline{0.676} & \underline{0.213} & 10.2K & \textbf{0.677} & 0.758 & \underline{0.153} & \underline{148K} & \textbf{0.546} & \underline{79.1K} \\
        0.5 & 0.10
        & 0.778 & \textbf{0.680} & \textbf{0.219} & \underline{8.20K} & \underline{0.620} & 0.742 & 0.141 & 165K & \underline{0.530} & 86.6K \\
        0.5 & 0.20
        & \underline{0.781} & 0.661 & 0.195 & \textbf{7.91K} & 0.581 & 0.756 & 0.138 & 151K & 0.519 & 79.5K \\
        1.0 & 0.05
        & 0.773 & 0.649 & 0.180 & 17.4K & 0.478 & \underline{0.760} & 0.141 & 150K & 0.497 & 83.7K \\
    \bottomrule
    \end{tabular}
    }
    \caption{Ablation study on the hyperparameters in total reward on Memalpha and MemoryAgentBench. \textbf{Bold} and \underline{underline} numbers indicate the best and second performance among evaluated methods. The \textbf{Len.} denotes the average memory length per sample in thousands of tokens.}
    \label{tab:ab_rewards}
\end{table*}

%% file: section/figures/ablation_beta.tex
\definecolor{red}{RGB}{172,21,28}
\definecolor{blue}{RGB}{39,89,167}

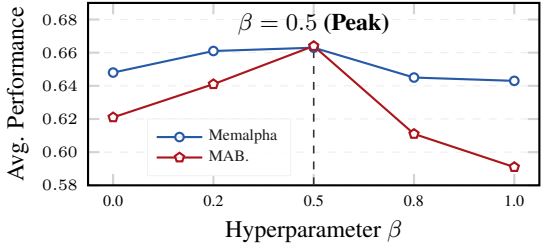
\begin{figure}[t]
    \centering
    \begin{tikzpicture}
    \begin{axis}[
        width=0.47\textwidth,
        height=0.25\textwidth,
        ymajorgrids=true,
        grid style={dashed, gray!40},
        axis line style={thick},
        tick style={thick},
        xmin=300, xmax=2100,
        ymin=0.58, ymax=0.69,
        xtick={400,800,1200,1600,2000,2400},
        xticklabels={0.0, 0.2, 0.5, 0.8, 1.0},
        ytick={0.58, 0.60, 0.62, 0.64, 0.66, 0.68, 0.70},
        yticklabel style={/pgf/number format/precision=2, /pgf/number format/fixed zerofill, font=\tiny},
        xticklabel style={font=\tiny},
        xlabel={Hyperparameter $\beta$},
        ylabel={Avg. Performance},
        xlabel style={font=\small, yshift=0.5em},
        ylabel style={font=\small, yshift=-0.8em},
        legend style={
            at={(0.45,0.05)}, 
            anchor=south east, 
            font=\tiny, 
            fill opacity=0.9, 
            draw=gray!30,
            cells={anchor=west},
            row sep=-2pt
        }
    ]
        \addplot[
            blue, 
            mark=*, 
            mark size=1.6pt, 
            thick, 
            mark options={fill=white, draw=blue, line width=0.8pt}
        ] coordinates {(400,0.648) (800,0.661) (1200,0.663) (1600,0.645) (2000,0.643)};
        \addlegendentry{Memalpha}
        \addplot[
            red, 
            mark=pentagon*, 
            mark size=1.8pt, 
            thick, 
            mark options={fill=white, draw=red, line width=0.8pt}
        ] coordinates {(400,0.621) (800,0.641) (1200,0.664) (1600,0.611) (2000,0.591)};
        \addlegendentry{MAB.}
        \draw[dashed, black!80, line width=0.6pt] (axis cs:1200, 0.58) -- (axis cs:1200, 0.664);
        \node[black, font=\small\bfseries, anchor=south] at (axis cs:1200, 0.665) {$\beta=0.5$ (Peak)};
    \end{axis}
    \end{tikzpicture}
    \vspace{-3mm}
    \caption{Ablation study on the hyperparameter $\beta$ in Evidence-Anchored Reward Attribution on Memalpha and MemoryAgentBench (\textbf{MAB.})}
    \label{fig:ablation_beta}
\end{figure}

%% file: section/figures/ablation_manager_reasoning_memalpha.tex
\definecolor{myorange}{RGB}{236, 154, 116}
\definecolor{myteal}{RGB}{84, 168, 178}

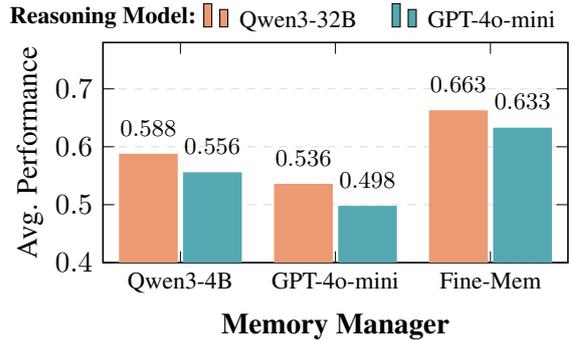
\begin{figure}[t]
    \centering
    \begin{tikzpicture}
        \pgfplotsset{
            empty legend/.style={draw=none, fill=none},
            myboxedstyle/.style={
                ybar,
                bar width=22pt,
                width=1.0\linewidth,
                height=4.5cm,
                ymin=0.40, ymax=0.78,
                enlarge x limits=0.25,
                ymajorgrids=true,
                grid style={dashed, gray!80},
                axis line style={draw=black, line width=0.6pt},
                tick align=inside,
                tick style={draw=black},
                ylabel style={font=\small},
                ylabel shift = -3pt,
                ylabel near ticks,
                xtick=data,
                xticklabel style={font=\small, align=center},
                nodes near coords,
                nodes near coords style={
                    font=\small\bfseries,
                    color=black,
                    /pgf/number format/fixed,
                    /pgf/number format/precision=3,
                    yshift=3pt
                },
            }
        }

        \node at (0, 3.25) [font=\small\bfseries] {Reasoning Model:};
        \begin{axis}[
            name=plot1,
            myboxedstyle,
            ylabel={Avg. Performance},
            symbolic x coords={Qwen3-4B, GPT-4o-mini, Fine-Mem},
            xlabel={\textbf{Memory Manager}},
            xlabel style={yshift= -1em, font=\tiny},
            xlabel near ticks,
            legend style={
                at={(0.6,1.00)},
                anchor=south,
                legend columns=-1,
                column sep=0.2em,
                draw=none,
                fill=none,
                /tikz/every even column/.append style={column sep=1 em},
                font=\small,
            },
            legend cell align={left},
        ]
            \addplot[fill=myorange, draw=none] coordinates {(Qwen3-4B, 0.588) (GPT-4o-mini, 0.536) (Fine-Mem, 0.663)};
            \addlegendentry{Qwen3-32B}
            
            \addplot[fill=myteal, draw=none] coordinates {(Qwen3-4B, 0.556) (GPT-4o-mini, 0.498) (Fine-Mem, 0.633)};
            \addlegendentry{GPT-4o-mini}
        \end{axis}
    \end{tikzpicture}
    
    \caption{Performance comparison of different Memory Managers combined with varying Reasoning Models on the Memalpha dataset.}
    \label{fig:ab_manager_reasoning_memalpha}
\end{figure}

%% file: section/tables/ablation_models.tex
\definecolor{forestgreen}{rgb}{0.13, 0.55, 0.13}

\begin{table}[t]
\centering
\small
\resizebox{\columnwidth}{!}{
    \begin{tabular}{llcccc}
    \toprule
    \textbf{Model} & \textbf{Methods} & \textbf{Memalpha} & \textbf{MAB.} & \textbf{Avg.} $\uparrow$ & \textbf{Len.} $\downarrow$ \\ \midrule
    
    \multirow{2}{*}{Qwen3-4B} & Base  & 0.588 & 0.615 & 0.601 & 61.4K \\
     & Fine-Mem & 0.663 & 0.664 & 0.663 \textcolor{forestgreen}{$\uparrow$}  & 79.1K \\ \cmidrule{1-6}
    
    \multirow{2}{*}{Qwen3-1.7B} & Base & 0.482 & 0.500 & 0.491 & 42.5K \\
     & Fine-Mem & 0.623 & 0.626 & 0.625 \textcolor{forestgreen}{$\uparrow$}  & 73.7K \\ \cmidrule{1-6}
    
    \multirow{2}{*}{Llama3.2-3B} & Base & 0.630 & 0.643 & 0.637 & 69.6K \\
     & Fine-Mem & 0.649  & 0.657  & 0.653 \textcolor{forestgreen}{$\uparrow$}  & 82.1K \\ \bottomrule
    \end{tabular}
}
    \caption{Performance of different models with Fine-Mem on Memalpha and MemoryAgentBench (\textbf{MAB.}). }
    \label{tab:ablation_models}
\end{table}

%% file: section/Related_work.tex
\paragraph{Memory-Augmented LLM Agents.}
Effective memory mechanisms are crucial for LLM agents to maintain coherent reasoning in long-horizon scenarios \citep{zhang2025survey, wang2024survey, du2025rethinking}. 
Early approaches \citep{modarressi2023ret, wang2023enhancing}, including MemGPT \citep{packer2023memgpt} and MemoryBank \citep{zhong2024memorybank}, adopted retrieval-augmented paradigms that offloaded interaction history to external memory. 
Building on this foundation, subsequent work introduced structural optimizations. Frameworks like MemTree \citep{rezazadeh2024isolated} and Zep \citep{rasmussen2025zep} organize context into hierarchical or temporal structures, while graph-based architectures \citep{chhikara2025mem0, gutierrez2025rag} explicitly model relational dependencies between memory nodes.
More recently, methods like MIRIX \citep{wang2025mirix}, and A-Mem \citep{xu2025mem} have shifted toward autonomous update schemes, where agents actively compress and prune information to optimize the retention-efficiency trade-off \citep{fang2025lightmem}, rather than passively accumulating history.
However, these methods depend on rigid workflows with strong LLMs, resulting in high overhead and low scalability \citep{hu2025evaluating}.

\paragraph{Reinforcement Learning for Memory Agent.}
To enhance the adaptability of memory systems, recent research has increasingly integrated RL with LLMs \citep{zhang2023rember, zhang2025learn, hu2025memory}, reframing memory management as a learnable policy rather than a static, rule-based mechanism. 
One prominent research direction focuses on training agents to autonomously manage their intrinsic working context. 
Approaches such as MEM1 \citep{zhou2025mem1} and MemAgent \citep{yu2507memagent} leverage RL to actively compress and reorganize information, thereby empowering agents to effectively navigate long-horizon dependencies. Conversely, a parallel line of inquiry adopts a decoupled architecture by introducing a dedicated memory controller. Systems like Memory-R1 \citep{yan2025memory} and \memalpha \citep{wang2025mem} employ a specialized manager to direct storage and retrieval operations independently of the reasoning agent.
Despite advancements, robust memory agents remain challenging due to sparse rewards and inefficient credit assignment in long-horizon tasks.

%% file: section/Conclusion.tex
This paper presents Fine-Mem, a unified reinforcement learning framework established for long-horizon memory management. 
Through the integration of the newly proposed Chunk-level Step Reward and Evidence-Anchored Reward Attribution, the approach effectively mitigates the challenges of reward sparsity and delayed credit assignment. 
Empirical evaluations on comprehensive benchmarks indicate that Fine-Mem consistently outperforms strong baselines.
Further ablation studies and hyperparameter analyses validate the efficacy and stability of Fine-Mem. 
Finally, experiments across diverse reasoning models and manager backbones validate the framework's strong generalization capabilities in various settings.

%% file: section/Limitations.tex
This work exhibits several limitations worth noting.
\textbf{Firstly}, our retrieval mechanism relies on BM25, despite being a simple and effective baseline for long-term memory, which limits the capture of deeper semantic relationships. Future research will explore richer alternatives, such as dense vector or graph-based retrieval.
\textbf{Secondly}, this work exclusively focuses on memory management within the textual mode.
We aim to extend our approach to support multimodal memory in future work, thereby broadening its real-world utility. 
\textbf{Thirdly}, although the decoupled training strategy effectively optimizes the memory manager, it does not support the enhancement of the reasoning model. 
Future research could investigate a co-evolutionary paradigm to achieve the synergistic optimization of both the manager and the reasoning agent.

%% file: section/Appendix.tex
\newpage
\section{Details of Memory Operations}
\label{app:operation_definitions}
\begin{figure}[h]
\centering
\includegraphics[clip, width=1.0\linewidth]{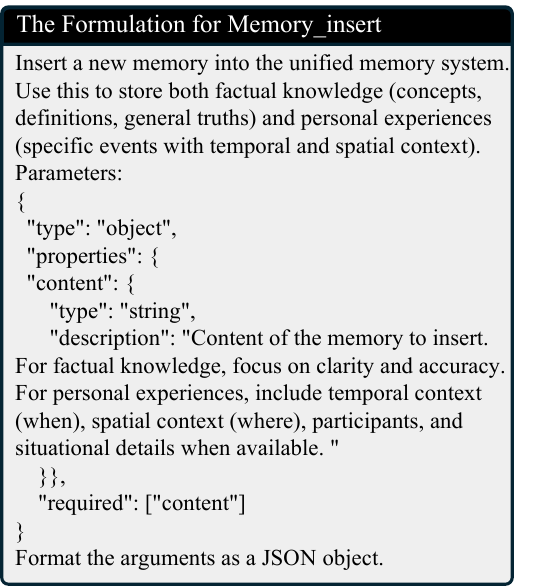}
\caption{JSON schema for the INSERT operation.}
\label{fig:memory_insert}
\end{figure}

\begin{figure}[t]
\centering
\includegraphics[clip, width=1.0\linewidth]{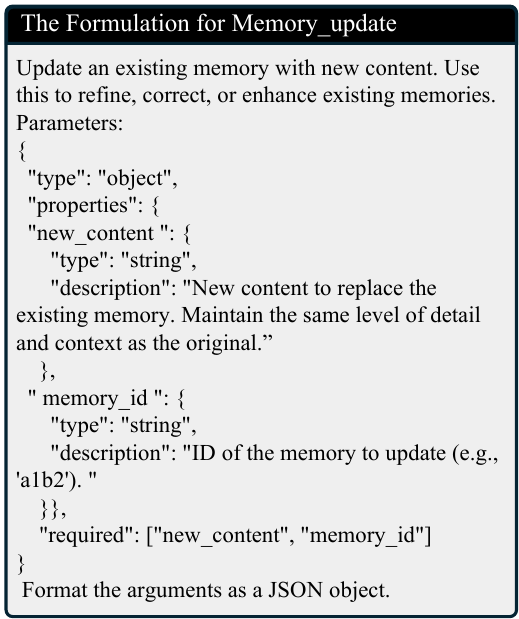}
\caption{JSON schema for the UPDATE operation.}
\label{fig:memory_update}
\end{figure}

\begin{figure}[t]
\centering
\includegraphics[clip, width=1.0\linewidth]{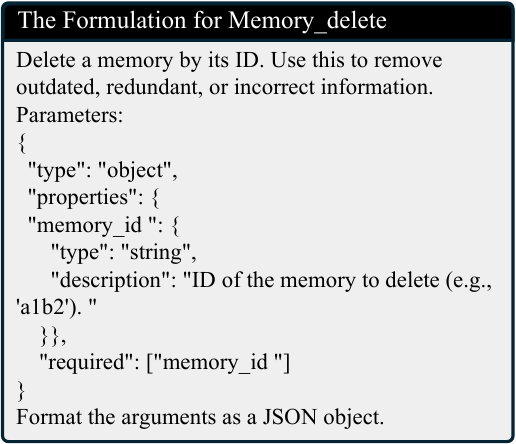}
\caption{JSON schema for the DELETE operation.}
\label{fig:memory_delete}
\end{figure}

\begin{figure*}[t]
\centering
\includegraphics[clip, width=1.0\linewidth]{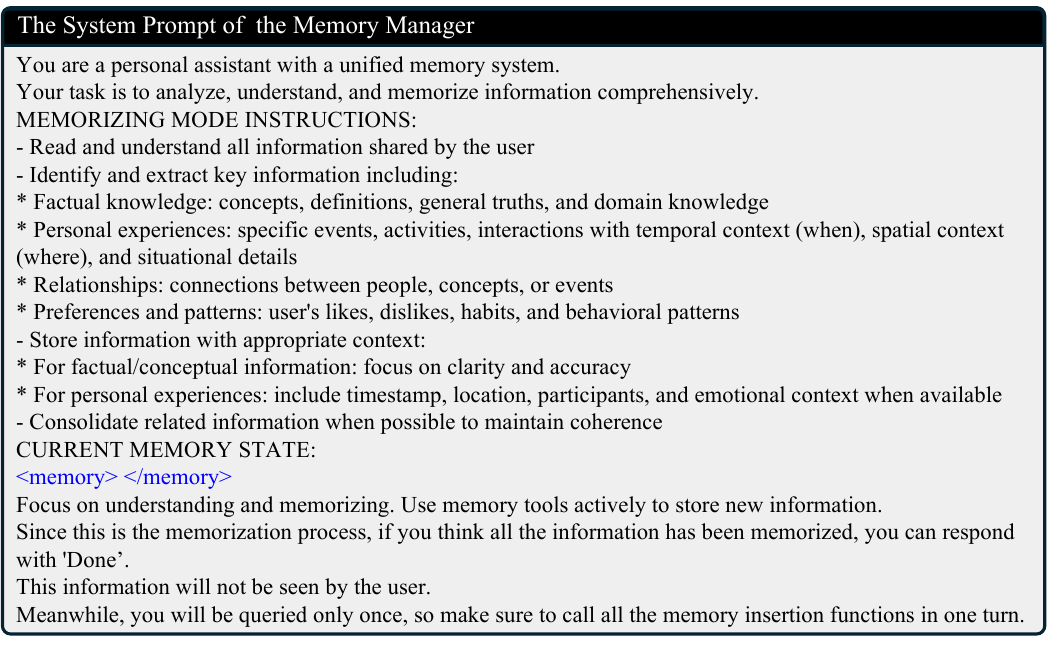}
\caption{System prompt for the memory manager with JSON-based function-call outputs.}
\label{fig:manager_prompt}
\end{figure*}

Consistent with prior work \citep{wang2025mem,yan2025memory}, we define the operation space of the memory manager as {INSERT, UPDATE, DELETE, SKIP}.
To align the manager’s behavior with these operations, we design a system prompt that constrains the model to produce outputs in a JSON-based function-call format, as illustrated in Figure \ref{fig:manager_prompt}.
When the SKIP operation is selected, the model is instructed to return the token “done”.
For the INSERT, UPDATE, and DELETE operations, the prompt enforces structured JSON schemas that explicitly specify the operation type, target object, and corresponding content, as detailed in Figures \ref{fig:memory_insert}, \ref{fig:memory_update}, \ref{fig:memory_delete}, respectively.

\section{Details of Reward Function}
\label{app:reward_fun}

\subsection{Chunk-Level QA Construction}
\label{app:chunk_qa}
To generate high-quality, verifiable supervision signals for RL, we design a multi-stage data engineering pipeline, as summarized in Algorithm \ref{alg:qa_construction}.

\begin{figure*}[t]
\centering
\includegraphics[clip, width=1.0\linewidth]{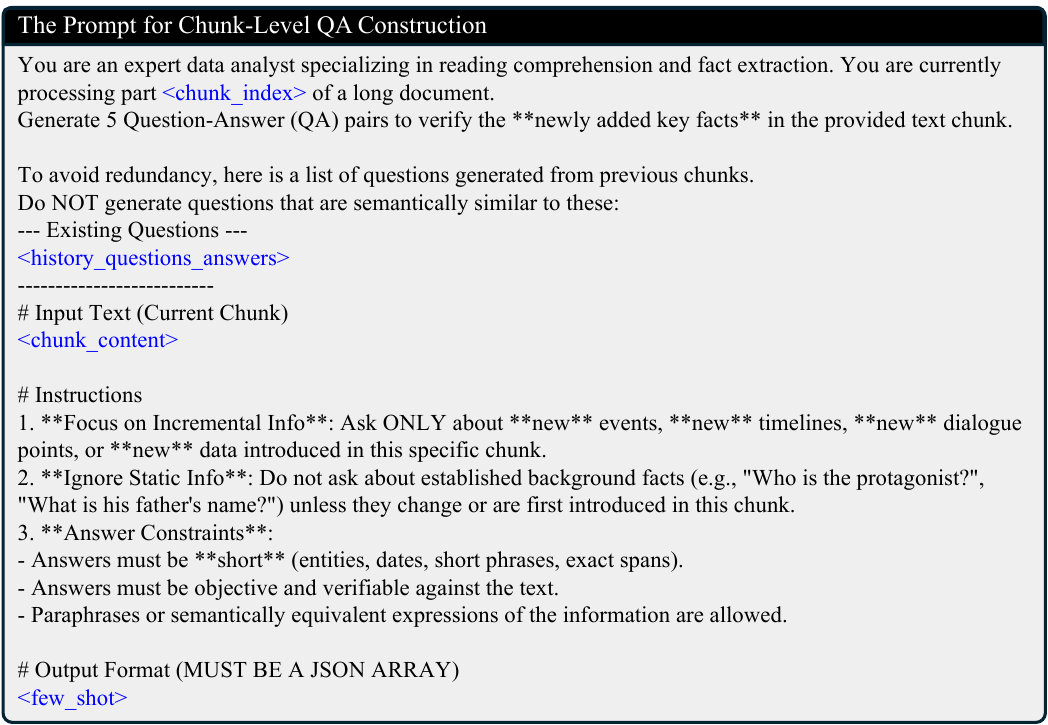}
\caption{Prompt for Chunk-Level QA Construction.}
\label{fig:chunk_qa_prompt}
\end{figure*}

For each chunk in an instance, we first prompt \texttt{GPT-4o-mini} to extract key information and convert it into QA pairs. 
The prompt is carefully crafted to encourage the generation of concise, factoid-style answers (e.g., entities, dates), which substantially reduces ambiguity in automated evaluation compared to open-ended generation. 
Moreover, to promote the reasoning agent's ability to generalize in downstream tasks based on the memory, we allow paraphrases of the content within the QA pairs, as detailed in Figures \ref{fig:chunk_qa_prompt}. 

Crucially, we incorporate a model-in-the-loop verification step to ensure data quality. 
Specifically, we employ \texttt{Qwen3-32B} as a verifier, asking it to answer each generated question using only the content of the corresponding chunk. 
A QA pair is considered valid only if the verifier can correctly derive the answer from the source text. 
This procedure ensures that the resulting dense rewards promote the extraction of information genuinely accessible from the text.

After removing duplicates with historical questions, we maintain a fixed budget of five QA pairs per chunk. 
This offline pre-computation avoids the latency of real-time LLM evaluation, supporting efficient and scalable training.

\input{section/tables/chunk_qa_construct}

\subsection{Proof of EARA Reward}
\label{app:proof_eara}

Consider a single rollout consisting of $T$ memory update steps and $n$ global QA pairs.
For the $j$-th question, let $M_j$ denote the set of retrieved memory items, and let $\phi(m)$ map a memory item $m$ to the update step at which it was generated. 

Let $s_j$ be the score associated with the $j$-th question. We define the \emph{Normalized Evidence Contribution (NEC)} for step $t$ as:
\begin{equation}
\label{eq:nec_def}
N_t = \sum_{j=1}^{n} \sum_{\substack{m \in M_j \\ \phi(m)=t}} \frac{s_j}{|M_j| \cdot n}.
\end{equation}
This quantity aggregates the evidence-based credit assigned to step $t$ from all questions, where each question's score is evenly distributed over its retrieved memory items and normalized by the total number of questions. 

Summing $N_t$ over all update steps yields:
\begin{equation}
\begin{aligned}
\sum_{t=1}^{T} N_t &= \sum_{t=1}^{T} \sum_{j=1}^{n} \sum_{\substack{m \in M_j \\ \phi(m)=t}} \frac{s_j}{|M_j| \cdot n} \\
&= \sum_{j=1}^{n} \frac{s_j}{n} \sum_{t=1}^{T} \sum_{\substack{m \in M_j \\ \phi(m)=t}} \frac{1}{|M_j|} \\
&= \sum_{j=1}^{n} \frac{s_j}{n} \sum_{m \in M_j} \frac{1}{|M_j|} \\
&= \frac{1}{n} \sum_{j=1}^{n} s_j.
\end{aligned}
\end{equation}
Since $\sum_{m \in M_j} \frac{1}{|M_j|} = 1$ for $j$-th question, and the rollout-level reward is defined as $r_{\textit{global}} = \frac{1}{n} \sum_{j=1}^{n} s_j$, it follows that:
\begin{equation}
\sum_{t=1}^{T} N_t = r_{\textit{}{global}}.
\end{equation}

EARA assigns the following reward to each update step $t$:
\begin{equation}
\label{eq:eara_reward}
r_{\textit{EARA}}^{(t)} = (1-\beta)\frac{r_{\textit{global}}}{T} + \beta N_t,
\end{equation}
where $\beta \in [0,1]$ controls the trade-off between uniform and evidence-based credit assignment. 

Summing the step-level rewards over the entire rollout gives:
\begin{equation}
\begin{aligned}
\sum_{t=1}^{T} r_{\textit{EARA}}^{(t)} &= (1-\beta){r_{\textit{global}}} \sum_{t=1}^{T} \frac{1}{T} + \beta \sum_{t=1}^{T} N_t \\
&= (1-\beta) r_{\textit{global}} + \beta r_{\textit{global}} \\
&= r_{\textit{global}}.
\end{aligned}
\end{equation}
Therefore, EARA achieves exact reward conservation, redistributing the rollout-level reward to step-level rewards without altering the total training signal. \hfill $\square$

\subsection{Hybrid Reward}
\label{app:hybrid_reward}
In this section, we introduce the mathematical formulations of the four components constituting our hybrid reward function: EARA for global QA Reward ($r_\textit{EARA}$), Chunk-Level Step Reward ($r_\textit{chunk}$), Formatting Validity Reward ($r_\textit{fmt}$), and Compression Efficiency Reward ($r_\textit{comp}$), referring to \memalpha. 
We define the memory state at time step $t$ as $\mathcal{M}_t$, and the manager's operations applied to the incoming chunk $c_t$ as $P_t$.

\paragraph{EARA for Global QA Reward}
This reward measures the long-term consistency of the memory system with respect to downstream task performance.
Given a global validation set $\mathcal{Q}{\text{glob}} = \{(q_i, y_i)\}_{i=1}^{N}$ derived from downstream tasks that require long-range reasoning, the reward is evaluated only at the final memory state $\mathcal{M}_T$.
For each query $q_i$, the reasoning agent $\pi_{\text{reason}}$ generates an answer conditioned on the final memory $\mathcal{M}_T$, which is denoted as $\pi_{\text{reason}}(\cdot \mid \mathcal{M}_T, q_i)$. The global QA reward is defined as:
\begin{equation}
r_{\textit{global}} = \frac{1}{N} \sum_{i=1}^{N}
\mathbb{I}\left[
\pi_{\text{reason}}(\cdot \mid \mathcal{M}_T, q_i), y_i \right]
\end{equation}
where $\mathbb{I}[\cdot]$ denotes an indicator function that returns $1$ if the prediction produced by the reasoning agent $\pi_{\text{reason}}$ is judged correct under the evaluation metric of the corresponding downstream task, and $0$ otherwise.
To enable fine-grained learning, we adopt \textit{Evidence-Anchored Reward Attribution (EARA)}, which redistributes the global reward across individual memory construction steps ($r_\textit{EARA}^{(t)}$) according to their contribution to successful reasoning outcomes, as shown in section \ref{meth:EARA}.

\paragraph{Chunk-level Step Reward}
To mitigate the supervision sparsity inherent in long-horizon tasks, we introduce a chunk-level step reward focused on local information retention. 
For each incoming chunk $c_t$, we automatically construct a set of local QA pairs $\mathcal{Q}_{local}^{(t)} = \{(q_j^{(t)}, y_j^{(t)})\}_{j=1}^{K}$ derived solely from the content of $c_t$. 
The reward at step $t$ verifies whether the updated memory $\mathcal{M}_t$ effectively retains key information based on the reasoning agent $\pi_{\text{reason}}$:
\begin{equation}
r_\textit{chunk}^{(t)} = \frac{1}{K} \sum_{j=1}^{K} \mathbb{I}\left[\pi_{\text{reason}}(\cdot | \mathcal{M}_t, q_j^{(t)}), y_j^{(t)} \right]
\end{equation}
This component incentivizes the agent to maximize immediate information precision, penalizing operations that discard essential details before global evaluation is possible.

\paragraph{Formatting Validity Reward}
To guarantee the operational integrity of the system during deployment, we enforce strict adherence to the defined schema. 
Let $\mathcal{P}_{valid}$ denote the set of permissible operation definitions and syntactic rules. 
For a sequence of operations $P_t = \{p_{t,1}, p_{t,2}, \dots, p_{t, M}\}$ generated at step $t$, the formatting reward is calculated as the average validity of the individual operations:
\begin{equation}
\begin{split}
    r_\textit{fmt}^{(t)} &= \frac{1}{M} \sum_{i=1}^M v(p_{t,i}), \\
    \text{where } v(p_{t,i}) &= \begin{cases} 
    1 & \text{if } p_{t,i} \in \mathcal{P}_{valid} \\ 
    0 & \text{otherwise} 
    \end{cases}
\end{split}
\end{equation}
This constraint ensures the policy learns to generate structurally sound memory updates that align with the environment's interface requirements.

\paragraph{Compression Efficiency Reward}
This reward regulates the trade-off between retention and storage, preventing memory size from scaling linearly with the input stream. 
Let $L(\cdot)$ denote the token length function. 
We define compression efficiency by comparing the final memory against the cumulative input size:
\begin{equation}
r_\textit{comp} = 1 - \frac{L(\mathcal{M}_T)}{\sum_{t=1}^{T} L(c_t)}
\end{equation}
This objective drives the policy towards succinct representations (e.g., abstraction or summarization) rather than copying.

Consequently, the total hybird reward $r_t$ for the sequence of operations $P_{t}$ is defined as a weighted linear combination:
\begin{equation}
    r_t = r_\textit{EARA}^{(t)} + r_\textit{fmt}^{(t)} + w_1 r_\textit{chunk}^{(t)} + w_2 r_\textit{comp}
\end{equation}
where $w_1$ and $w_2$ denote hyperparameters that balance the contributions of these distinct reward components.

\section{Detailed GRPO Formulation}
\label{appendix:grpo}
In this section, we provide the complete formulation of the GRPO objective used to train the \textit{Memory Manager}. 

For each input chunk $c_t$ at step $t$, we sample a group of $G$ operations $\{P_{t,1}, \dots, P_{t,G}\}$ from the old policy $\pi_{\text{old}}$. 
For each sampled sequence $P_{t,j}$, we compute a hybrid reward $r_{t,j}$, which evaluates the overall quality of the operations performed at step $t$. 
The group-relative advantage for the $j$-th sequence is then obtained by standardizing these rewards:
\begin{align}
A_{t,j} = \frac{r_{t,j} - \mu_{\text{group}}^{(t)}}{\sigma_{\text{group}}^{(t)} + \epsilon},
\end{align}
where $\mu_{\text{group}}^{(t)}$ and $\sigma_{\text{group}}^{(t)}$ denote the mean and standard deviation of rewards across the group of operation sequences at step $t$, respectively.

The final GRPO objective is defined as:
\begin{align}
\mathcal{J}(\theta)
& = \mathbb{E} \Bigg[
    \frac{1}{G} \sum_{t=1}^G 
    \frac{1}{|P_t|} \sum_{i=1}^{|P_t|} \min \bigg( 
    \rho_t(\theta),    \nonumber
\\
    &\quad \text{clip}\big(\rho_t(\theta), 1-\varepsilon, 1+\varepsilon\big)\,
    \bigg)A_t
\Bigg] \nonumber, \\[6pt]
\text{where} \quad
\rho_t(\theta) 
&= \frac{
    \pi_\theta(P_t \mid \mathcal{M}_{t-1}, c_t)
}{
    \pi_{\text{old}}(P_t \mid \mathcal{M}_{t-1}, c_t)
}, \nonumber \\
P_t &= \{P_{t,1}, \dots, P_{t,G}\}.
\end{align}
Following standard practices, we employ the clipping to constrain the policy update and integrate a KL-divergence term into the loss function.

\section{Details of Datasets}
\label{app:datasets}
In this section, we provide an overview of the training dataset and evaluation datasets. 
Following \memalpha \citep{wang2025mem}, we foucs on memory agents through three core competencies: \textbf{Accurate Retrieval} (AR), which measures the agent’s ability to store factual information and retrieve it precisely from memory faithfully; \textbf{Test-Time Learning} (TTL), which assesses the capacity to acquire and apply new rules or patterns introduced dynamically during interaction; \textbf{Long-Range Understanding} (LRU), which evaluates the ability to maintain global context over long horizons and synthesize a coherent understanding.

\subsection{Training Dataset}
\label{app:train_dataset}
We construct a training corpus based on the data used in \memalpha, further augmented with a newly constructed chunk-level QA dataset. 
The combined datasets form a multi-task mixture aligned with three core memory-related competencies: Accurate Retrieval (AR), Test-Time Learning (TTL), and Long-Range Understanding (LRU). 
(1) The AR category includes SQuAD \citep{squad}, HotpotQA \citep{hotpotqa}, PerLTQA \citep{perltqa}, and LME-Train \citep{longmemeval}; 
(2) The TTL category adopts classification-oriented datasets such as PubMed-RCT \citep{pubmed}, NLU, and TREC-Coarse \citep{hu2025evaluating}; 
(3) The LRU category employs the BookSum dataset \citep{booksum}, segmented into sequential chunks. 
All datasets are presented in a sequential format, where information is organized incrementally across chunks to facilitate memory accumulation.  
The resulting statistics are summarized in Table~\ref{tab:training_data_stats}.

\input{section/tables/train_dataset_stats}

\subsection{Evaluation Dataset}
\label{app:eval_dataset}
Aligned with the \memalpha, our evaluation assesses both in-distribution (ID) and out-of-distribution (OOD) performance.
The ID evaluation utilizes the \memalpha validation set, comprising 468 instances across 7 subsets (excluding LME-Train used in training);
The OOD evaluation employs MemoryAgentBench~\citep{hu2025evaluating}, which features 112 instances across 9 datasets characterized by significantly longer context lengths.
Specifically for MemoryAgentBench, the datasets are aligned with three core competencies:
(1) The AR category includes RULER-QA1 and RULER-QA2 \citep{ruler}, which test single-document and multi-document question-answering capabilities, respectively, as well as LME(S*) \citep{wu2024longmemeval} for evaluating precise information retrieval;
(2) The TTL category adopts five classification datasets covering varying granularities: TREC-Coarse and TREC-Fine \citep{li2002learning}, NLU \citep{liu2021benchmarking}, CLINIC150 \citep{larson2019evaluation}, and Banking77 \citep{casanueva2020efficient};
(3) The LRU category uses the InfBench-Sum dataset \citep{zhang2024bench}, which requires high-level summarization from full-length novels.

\input{section/tables/eval_dataset_stats}

We assess the performance across memory capabilities using five complementary metrics according to the nature of each task:
\noindent(1) \textbf{SubEM (Subsection Exact Match)}: Measures strict retrieval recall by checking if the ground-truth answer appears verbatim in the responses.
\noindent(2) \textbf{EM (Exact Match)}: Evaluates classification or rigid-format tasks by computing the percentage of exact matches between predicted and ground-truth outputs.
\noindent(3) \textbf{Source-based}: Verifies whether responses are correctly derived from specific memory chunks, ensuring factual support from the source rather than hallucination.
\noindent(4) \textbf{LLM Judge (LLM-J)}: Uses Qwen3-32B to semantically assess open-ended outputs against references, focusing on meaning rather than exact string matches.
\noindent(5) \textbf{KW Hit (Keyword Hit)}: Measures the recall of key entities, events, or concepts from the reference in the generated output, reflecting information coverage and synthesis.

The statistics for all evaluation datasets are summarized in Table~\ref{tab:eval_dataset_stats}.

\section{Details of Implementation}
\label{app:implementation}

\subsection{Baselines}
\label{app:baselines}
In this section, we describe the detailed implementation of all baseline methods used in our evaluation, which can be grouped into three categories, comprising seven approaches.
Unless otherwise specified, we adopt \textbf{Qwen3-32B} with a \textbf{32k-token context window} as the reasoning agent for all baselines and uniformly deploy it in a \textbf{non-thinking} inference mode.

\begin{itemize}[leftmargin=*]
\item \textbf{Non-Constructive Memory}.  

(1) \textbf{Long-Context}.  
The reasoning agent directly processes the available context using its maximum context window. 
When the total length of accumulated chunks exceeds 32k tokens, only the most recent 32k tokens are retained.

(2) \textbf{RAG-Top2}.  
A retrieval-augmented baseline based on BM25, where the query is used to retrieve the top two most relevant chunks from the historical context, which are then provided to the reasoning agent for answer generation.

\item \textbf{Workflow-Based Memory Systems}.  

(3) \textbf{A-Mem} \citep{xu2025mem}.  
A dynamic agentic memory system that incrementally creates, links, and updates structured memory units to support long-term and cross-session reasoning.
For a fair comparison, we replace its default embedding model (\texttt{all-MiniLM-L6-v2}) with \texttt{Qwen3-Embedding-0.6B}, which better accommodates long chunks and avoids performance degradation caused by aggressive truncation.

(4) \textbf{LightMem} \citep{fang2025lightmem}.  
A cognitively inspired memory framework that emphasizes efficiency through compressive sensory filtering, topic-aware consolidation, and decoupled sleep-time updates.
To accommodate this system, we adapt datasets with different task formats into a unified \textbf{dialogue-style input}, enabling proper compression and memory storage as required by the framework.

\item \textbf{RL-Based Memory Agents}.  

(5) \textbf{MemAgent} \citep{yu2507memagent}.  
An RL-trained agent that iteratively processes all available chunks under a task-specific instruction and forms an internal memory state, which is directly used to answer downstream questions.
In our experiments, we use the released model \texttt{BytedTsinghua-SIA/RL-MemoryAgent-14B} to construct and maintain the memory.

(6) \textbf{MEM1} \citep{zhou2025mem1}.  
An RL-based agent that maintains a single-paragraph memory representation, which is continuously retrieved and updated as new information becomes available.
During evaluation, we employ the released model \texttt{Mem-Lab/Qwen2.5-7B-RL-RAG-Q2-EM-Release} for memory construction.

(7) \textbf{\memalpha} \citep{wang2025mem}.  
A hierarchical memory system with a specialized memory manager trained using a GRPO-based objective to optimize memory operations.
\end{itemize}

Furthermore, we excluded the related method, Memory-R1 \citep{yan2025memory}, from our comparison as its code and data are not publicly available. 
It is worth noting that Memory-R1 relies solely on final-answer accuracy as a reward signal, a setting we replicate in our ablation using only the global QA reward to assess its effect in section \ref{sec:ab_study}.

\subsection{Fine-Mem}
\label{app:fine-mem}
Our method is implemented based on a modified version of the VERL framework~\citep{sheng2024hybridflow}. 
We utilize \texttt{Qwen3-4B} as the backbone for the memory manager and deploy a long-context \texttt{Qwen3-32B} via vLLM~\citep{kwon2023efficient} as the reasoning agent. 
To ensure generation stability, the temperature for the reasoning agent is set to $0.1$. 
Both the manager and the reasoning agent operate in a non-thinking mode, and BM25 is employed for retrieval. 
Consistent with the settings in \memalpha, we configure the learning rate to $1\times 10^{-6}$, the batch size to 32, and the number of rollouts per prompt (\texttt{grpo\_rollout\_n}) to 8. 
The reward weights are set as $w_1 = 0.5$, $w_2 =0.05$, and $\beta=0.5$. 
To accelerate the training process, we compute global rewards using a random 20\% subset of Global QA pairs for each instance, which has been proven to yield performance comparable to training on the full dataset. 
Regarding training, we observe that a single-layer memory structure converges significantly faster than multi-layer variants.
Consequently, we train the models for 2 epochs and utilize the final checkpoint. 
For a fair comparison, we report the results for \memalpha following the original setting (obtained by training for 85 steps on the full Global QA dataset).

\input{section/figures/ablation_manager_reasoning_memoryagentbench}

%% file: section/tables/chunk_qa_construct.tex
\begin{algorithm}[t]
\caption{Construction of Chunk-level QA Dataset}
\label{alg:qa_construction}
\begin{algorithmic}[1]
\Input Dataset $\mathcal{D}$, Teacher $\pi_{\text{tea}}$ (GPT-4o-mini), Verifier $\pi_{\text{ver}}$ (Qwen3-32B)
\Input Hyperparameter: Top-$K$ ($K=5$)
\Output Augmented Dataset with Chunk-level QA pairs $\mathcal{D}_{\text{chunk}}$

\State Initialize $\mathcal{D}_{\text{chunk}} \leftarrow \emptyset$

\For{each instance $I = \{c_1, c_2, \dots, c_T\}$ in $\mathcal{D}$}
    \State $H_{\text{qa}} \leftarrow \emptyset$; $I_{\text{chunk}} \leftarrow \emptyset$
    
    \For{each chunk $c_t$ in $I$}
        \State $\mathcal{Q}_{\text{raw}} \leftarrow \pi_{\text{tea}}(c_t)$ 
        
        \State $\mathcal{Q}_{\text{verified}} \leftarrow \emptyset$
        
        \For{$(q, a)$ in $\mathcal{Q}_{\text{raw}}$}
            \State $\hat{a} \leftarrow \pi_{\text{ver}}(q, c_t)$
            
            \If{$\text{Match}(\hat{a}, a)$ \textbf{and} $(q, a) \notin H_{\text{qa}}$}
                \State $\mathcal{Q}_{\text{verified}}.\text{add}((q, a))$
                \State $H_{\text{qa}}.\text{add}((q, a))$
            \EndIf
        \EndFor
        
        \State $\mathcal{G}_t \leftarrow \text{SelectTopK}(\mathcal{Q}_{\text{verified}}, K)$
        \State $I_{\text{chunk}}.\text{append}((c_t, \mathcal{G}_t))$
    \EndFor
    \State $\mathcal{D}_{\text{chunk}}.\text{add}(I_{\text{chunk}})$
\EndFor

\State \Return $\mathcal{D}_{\text{chunk}}$
\end{algorithmic}
\end{algorithm}

%% file: section/tables/train_dataset_stats.tex
\begin{table}[t]
\centering
\begin{tabular}{llrr}
\toprule
\textbf{Category} & \textbf{Dataset} & \textbf{Ins.} & \textbf{ChunkQA} \\
\midrule
\multirow{4}{*}{AR} 
 & SQuAD & 100 & 4975 \\
 & HotpotQA & 100 & 4840 \\
 & PerLTQA & 27 & 3140 \\
 & LME-Train & 50 & 3845 \\
\midrule
\multirow{3}{*}{TTL} 
 & NLU & 49 & 2450 \\
 & TREC-Coarse & 51 & 2550 \\
 & PubMed-RCT & 90 & 4500 \\
\midrule
LRU & BookSum & 100 & 3915 \\
\midrule
\multicolumn{2}{l}{\textbf{Total}} & \textbf{567} & \textbf{30215} \\
\bottomrule
\end{tabular}
\caption{Statistics of the training dataset. 
\textbf{AR}: Accurate Retrieval, 
\textbf{TTL}: Test-Time Learning, 
\textbf{LRU}: Long-Range Understanding. 
\textbf{Ins.} denotes the number of instances, and 
\textbf{ChunkQA} denotes the number of chunk-level QA .}
\label{tab:training_data_stats}
\end{table}

%% file: section/tables/eval_dataset_stats.tex
\begin{table*}[t]
\centering
\resizebox{\textwidth}{!}{
\begin{tabular}{clllrrrr}
\toprule
\multirow{2}{*}{\textbf{Category}} & \multirow{2}{*}{\textbf{Dist.}} & \multirow{2}{*}{\textbf{Dataset}} & \multirow{2}{*}{\textbf{Metric}} & \multicolumn{4}{c}{\textbf{Statistics}} \\
\cmidrule(lr){5-8}
 & & & & \textbf{Ins.} & \textbf{Ch/Ins} & \textbf{Tok/Ch} & \textbf{Q/Ins} \\
\midrule
\multirow{5}{*}{\textbf{AR}} 
 & \textit{ID} & SQuAD & SubEM & 30 & 10.0 & 1,057.0 & 96.8 \\
 & \textit{ID} & HotpotQA & SubEM & 219 & 9.2 & 1,051.6 & 17.0 \\
 & \textit{ID} & PerLTQA & SubEM & 4 & 23.0 & 567.8 & 100.0 \\
 \cdashline{2-8}
 & \textit{OOD} & LongMemEval & LLM-J & 5 & 218.6 & 1,591.4 & 60.0 \\
 & \textit{OOD} & RULER-QA1/2 & Source & 2 & 161.0 & 2,133.7 & 100.0 \\
\midrule
\multirow{8}{*}{\textbf{TTL}} 
 & \textit{ID} & NLU & EM & 20 & 10.0 & 606.2 & 100.0 \\
 & \textit{ID} & TREC-Coarse & EM & 20 & 10.0 & 390.2 & 100.0 \\
 & \textit{ID} & PubMed-RCT & EM & 10 & 10.0 & 1,673.3 & 100.0 \\
 \cdashline{2-8}
 & \textit{OOD} & Banking77 & Source & 1 & 111.0 & 1,150.3 & 100.0 \\
 & \textit{OOD} & Clinic150 & Source & 1 & 38.0 & 3,440.5 & 100.0 \\
 & \textit{OOD} & NLU & EM & 1 & 115.0 & 1,166.7 & 100.0 \\
 & \textit{OOD} & TREC-Coarse & EM & 1 & 111.0 & 1,114.6 & 100.0 \\
 & \textit{OOD} & TREC-Fine & EM & 1 & 108.0 & 1,163.3 & 100.0 \\
\midrule
\multirow{2}{*}{\textbf{LRU}} 
 & \textit{ID} & BookSum & KW Hit & 155 & 8.1 & 1,914.3 & 1.0 \\
 \cdashline{2-8}
 & \textit{OOD} & InfBench-Sum & Source & 100 & 88.9 & 2,034.1 & 1.0 \\
\bottomrule
\end{tabular}
}
\caption{Detailed statistics of the evaluation datasets. We categorize the datasets into three task types: AR, TTL, and LRU, covering both In-Distribution (ID) and Out-of-Distribution (OOD) scenarios. \textbf{Ins.}, \textbf{Ch/Ins}, \textbf{Tok/Ch}, and \textbf{Q/Ins} denote the number of instances, chunks per instance, tokens per chunk, and queries per instance, respectively.}
\label{tab:eval_dataset_stats}
\end{table*}

%% file: section/figures/ablation_manager_reasoning_memoryagentbench.tex
\definecolor{myorange}{RGB}{236, 154, 116}
\definecolor{myteal}{RGB}{84, 168, 178}

\begin{figure}[t]
    \centering
    \begin{tikzpicture}
        \pgfplotsset{
            empty legend/.style={draw=none, fill=none},
            myboxedstyle/.style={
                ybar,
                bar width=22pt,
                width=1.0\linewidth,
                height=5.0cm,
                ymin=0.40, ymax=0.78,
                enlarge x limits=0.25,
                ymajorgrids=true,
                grid style={dashed, gray!80},
                axis line style={draw=black, line width=0.6pt},
                tick align=inside,
                tick style={draw=black},
                ylabel style={font=\small},
                ylabel shift = -3pt,
                ylabel near ticks,
                xtick=data,
                xticklabel style={font=\small, align=center},
                nodes near coords,
                nodes near coords style={
                    font=\small\bfseries,
                    color=black,
                    /pgf/number format/fixed,
                    /pgf/number format/precision=3,
                    yshift=3pt
                },
            }
        }

        \node at (0, 3.90) [font=\small\bfseries] {Reasoning Model:};
        \begin{axis}[
            name=plot1,
            myboxedstyle,
            ylabel={Avg. Performance},
            symbolic x coords={Qwen3-4B, GPT-4o-mini, Fine-Mem},
            xlabel={\textbf{Memory Manager}},
            xlabel style={yshift=0.5em, font=\small},
            xlabel near ticks,
            legend style={
                at={(0.6,1.05)},
                anchor=south,
                legend columns=-1,
                column sep=0.2em,
                draw=none,
                fill=none,
                /tikz/every even column/.append style={column sep=1 em},
                font=\small,
            },
            legend cell align={left},
        ]
            \addplot[fill=myorange, draw=none] coordinates {(Qwen3-4B, 0.615) (GPT-4o-mini, 0.622) (Fine-Mem, 0.664)};
            \addlegendentry{Qwen3-32B}
            
            \addplot[fill=myteal, draw=none] coordinates {(Qwen3-4B, 0.601) (GPT-4o-mini, 0.624) (Fine-Mem, 0.655)};
            \addlegendentry{GPT-4o-mini}
        \end{axis}
    \end{tikzpicture}
    
    \caption{Performance comparison of different Memory Managers combined with varying Reasoning Models on MemoryAgentBench dataset.}
    \label{fig:ab_manager_reasoning_MAB}
\end{figure}
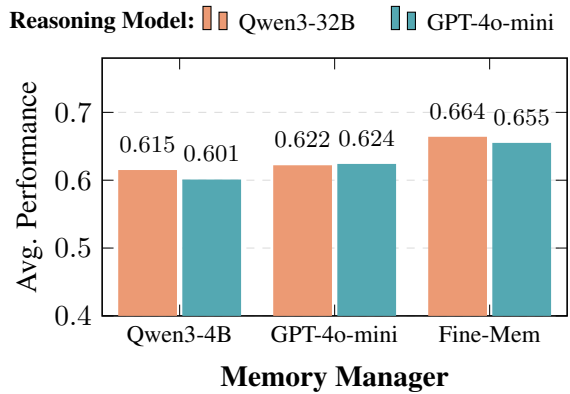